# Neural Network Applications in Earthquake Prediction (1994-2019): Meta-Analytic Insight on their Limitations

Arnaud Mignan[1,2,3*] and Marco Broccardo[3,4]

*Version: 24 September 2019*

[1] Institute of Geophysics, Swiss Federal Institute of Technology, Zurich, Switzerland

[2] Institute of Risk Analysis, Prediction and Management, Academy for Advanced Interdisciplinary Studies, Southern University of Science and Technology, Shenzhen, China

[3] Swiss Seismological Service, Zurich, Switzerland

[4] Institute of Structural Engineering, Swiss Federal Institute of Technology, Zurich, Switzerland

[*] arnaud.mignan@sed.ethz.ch

*Abstract:* In the last few years, deep learning has solved seemingly intractable problems, boosting the hope to find approximate solutions to problems that now are considered unsolvable. Earthquake prediction, the Grail of Seismology, is, in this context of continuous exciting discoveries, an obvious choice for deep learning exploration. We review the entire literature of artificial neural network (ANN) applications for earthquake prediction (77 articles, 1994-2019 period) and find two emerging trends: an increasing interest in this domain, and a complexification of ANN models over time, towards deep learning. Despite apparent positive results observed in this corpus, we demonstrate that simpler models seem to offer similar predictive powers, if not better ones. Due to the structured, tabulated nature of earthquake catalogues, and the limited number of features so far considered, simpler and more transparent machine learning models seem preferable at the present stage of research. Those baseline models follow first physical principles and are consistent with the known empirical laws of Statistical Seismology, which have minimal abilities to predict large earthquakes.

# 1 Introduction

Deep learning is rapidly rising as one of the most powerful go-to techniques not only in data science [1-2] but also for solving hard and intractable problems of Physics [3-5]. This is justified by the superior performance of deep learning in discovering complex patterns in very large datasets. One of the major advantages of artificial neural networks (ANNs), including deep neural networks (DNNs), is that, generally, there is no need for feature extraction and engineering, as data can be used directly to train the network with potentially great results. It comes as no surprise that machine learning at large—including deep learning—has become popular in Statistical Seismology [6-7] and gives fresh hope for earthquake prediction[1] [8-9]. This challenge has long been considered impossible [10], although the Seismological community has already gone through several cycles of optimism/pessimism over the past decades [11]. It seems that we have now entered a new phase of enthusiasm with machine learning-based earthquake forecasting [8-9,12]. Another boost of activity comes from the generation of improved earthquake catalogues based on convolutional neural networks (CNN) [13-14], although the potential use of those "Big Data catalogues" to improve earthquake predictability has yet to be investigated.

In the specific case of ANNs, designing a suitable architecture can be a highly iterative process, based on hyper-parameterization tuning and, sometimes (even if not highly recommended in this context) feature engineering. How do such choices affect, not only model performance, but physical interpretability? In view of the flexibility of ANNs and their black-box nature, can we miss critical scientific insights in the modelling process? We aim to answer these questions by developing upon the preliminary study of Mignan and Broccardo [15], first by expanding their original review to provide a comprehensive survey of the ANN-based earthquake prediction literature (section 2), and second by demonstrating that much simpler models, which can be related to first physical principles, may yield similar or better prediction performances (section 3).

While our arguments are supported by observations in Statistical Seismology, they can generalize to any other applied science domain where researchers may also be carried away by the apparent power of ANNs and in particular deep learning, a problem recently re-emphasized in the data science community [16-18]. We, however, clearly support the use of deep learning in computer vision applications based on unstructured data (e.g., seismic

---

[1] The term "prediction" is used in most of the papers considered in this survey, although it is often a probabilistic forecast that is offered.

waveform data). Pros and cons for Seismology are discussed in some concluding remarks (section 4).

## 2 ANN-based earthquake prediction, a literature survey (1994-2019)

"*The subject of Statistical Seismology aims to bridge the gap between physics-based models without statistics, and statistics-based models without physics*" [19]. This scientific domain can then be divided into two categories, with earthquakes as point sources (i.e. seismicity) and modeled as (non-)stationary stochastic point processes, or earthquakes as seismic waves radiating from finite sources. We are here concerned with the applicability of ANNs in seismicity analyses that pertain to earthquake forecasting and prediction; we will only briefly comment on ANN applications in seismic waveform analysis, in which case deep learning is far more pertinent due to the data being unstructured in contrast to tabulated in earthquake catalogues.

We developed a comprehensive corpus of 77 articles, spanning from 1994 to 2019, on the topic of ANN-based earthquake prediction. Although a few references may have been missed, the survey can be considered complete enough to investigate emerging trends via a meta-analysis [11,20-22]. The full database, `DB_EQpred_NeuralNets_v1.json`, is available at `github.com/amignan/hist_eq_pred/`.

Figure 1 shows the annual rate of publications over time, which indicates a progressive increase in the number of studies on this topic. Only in the past ten years did important papers emerge in terms of number of citations and journal impact factor [9,23-24]. We will divide this review into two sections, (*i*) the classical ANN literature, with two hidden layers maximum (section 2.1), and (*ii*) the very recent highly-parameterized deep learning trend, up to 6-hidden-layer DNN [9] and 3-convolutional-layer CNN [25] (section 2.2). It should be noted that our survey differs from previous reviews [26] by being more systematic and analytical, but less descriptive. Those studies are thus complementary to each other.

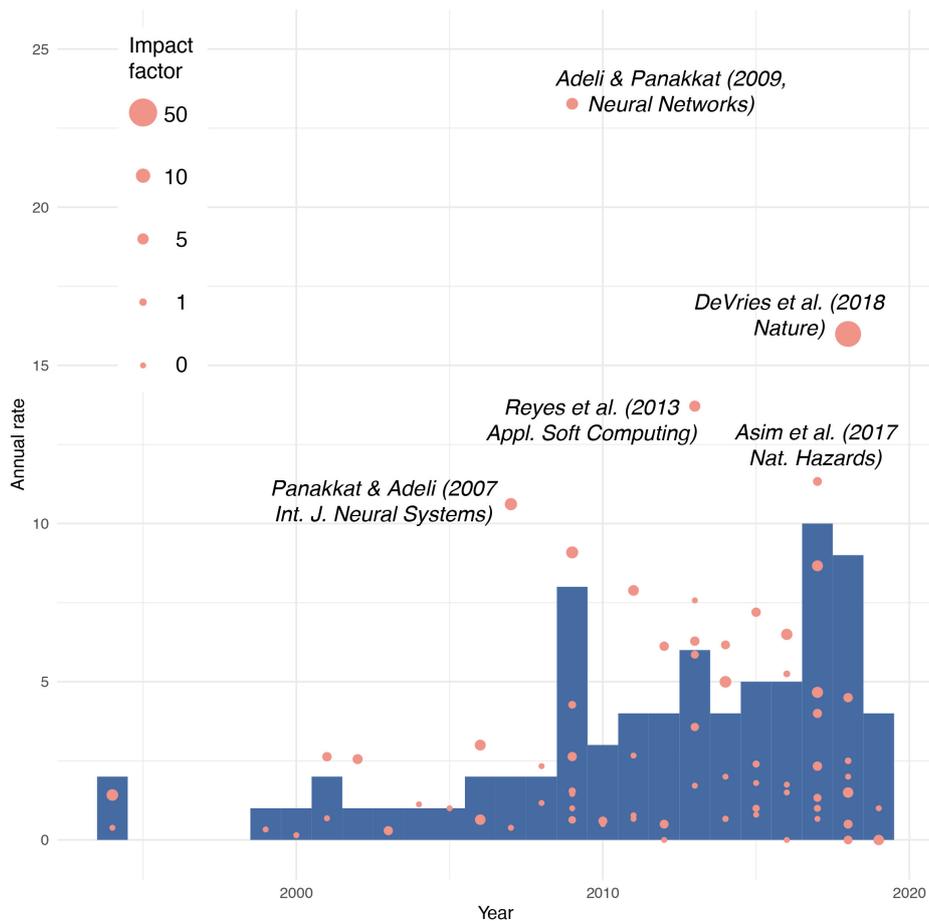

**Fig. 1** Annual rate of publications and citations over the period 1994-2019 for a comprehensive corpus of 77 articles on the topic of ANN-based earthquake prediction. Note the progressive increase in number of studies and in scientific impact, the latter being described by higher citation rate and journal impact factor

*2.1 Classical ANN-based earthquake prediction literature*

ANNs were introduced in Seismology as early as 1990 [27], only four years after the seminal back-propagation article of Rumelhart, Hinton & Williams [28]. The earliest attempts to apply ANNs to earthquake prediction date back, to the best of our knowledge, to 1994 [29-30]. Few studies followed in the next years [31]. The first deep neural network (DNN), with two hidden layers, was proposed in 2002 [32] and the first recurrent neural network (RNN) in 2007 [33]. Panakkat and Adeli [33] provided the first comprehensive work on ANN applications to earthquake prediction, comparing three types of neural networks: a radial basis function (RBF) neural network, a DNN and an RNN. The diversity of ANN architectures used for earthquake prediction is shown in Table 1.

**Table 1** ANN models for earthquake prediction based on structured earthquake catalogues

| Input layer* | Architecture† (cases/77) | Output layer |
|---|---|---|
| *Size (magnitude, energy):* mean, quantiles, seismicity law parameters | MLP (50.6%) DNN (16.9%) RBF (7.8%) ANFIS (3.9%) RNN (3.9%) LSTM (3.9%) ANN ensemble (3.9%) SOM (2.6%) PNN (2.6%) CNN (1.3%) NARX (1.3%) wavelet NN (1.3%) | *Mainshock size:* magnitude $m$, max($m$), $m$ threshold, felt intensity |
| *Location:* longitude, latitude, depth distribution, clustering | | *Location of mainshock:* longitude, latitude, depth |
| *Count in time:* seismicity rate, interevent time, aperiodicity, aftershock law parameters, financial market indices | | *Mainshock occurrence:* occurrence time, interevent time |
| *Non-seismicity:* slip rate, static stress, geo-electric, ionospheric, radon, *etc.* | | |

* Feature transformations may include feature splits, derivation, deviation from trend, exponentiation, principal component analysis, and/or declustering;
† MLP: multi-layer perceptron, DNN: deep feed-forward fully connected neural network, RBF: radial basis function neural network, ANFIS: adaptive neuro fuzzy inference system, RNN: recurrent neural network (excl. LSTM), LSTM: long short-term memory neural network, SOM: self-organizing map, PNN: probabilistic neural network, CNN: convolutional neural network, NARX: nonlinear autoregressive exogeneous model with neural network.

Most applications use time series data, with seismicity indicators estimated from discretized bins and used as ANN input units (a same approach can be used in space, with data discretized in geographic cells). The size of the input layer varies from 2 to 94 neurons with a median of 7 and mean of 10 (excluding the CNN case, see section 2.2.2). Note that although 75% of the corpus studies use seismicity as input, others use geo-electric (4%), ionospheric (4%) or other signals (such as radon or stress); we here focus on seismicity-based analyses where primary data consist of occurrence time, magnitude, longitude, latitude and depth vectors. The output units predict future mainshock characteristics, most often related to the event magnitude in a time or space-time window (mainshock occurrence time and location are more seldom predicted [34]). Their number is most often 1 (minimum, median and mean obtained from the corpus), corresponding to a binary classification (e.g. mainshock above threshold $m_{th}$ or not) or a regression (e.g. mainshock magnitude $m$ estimate). A list of the main features and outputs used in the corpus is given in Table 1.

Features are standard statistical metrics, such as *n*th-order moments or quantiles, seismicity-based metrics [33,35-36] or, rarely, metrics used in financial analysis [37]. So-called seismicity indicators are often the parameters of the Gutenberg-Richter (GR) law (Eq.

1a) [38] and/or of the Modified Omori law (MOL) (Eq. 1b) [39-40], which are the main empirical laws of Statistical Seismology:

$$\lambda(\geq m_{th}) = 10^{a-bm_{th}} \quad (1a)$$

$$\lambda(t, \geq m_{th}) = K(\geq m_{th})/(t+c)^p \quad (1b)$$

The GR law provides the rate of earthquakes above a certain magnitude $m_{th}$ depending on the overall seismic activity $a$ and magnitude ratio $b$; the MOL provides the rate of aftershocks occurring $t$ days after a mainshock depending on the number of produced aftershocks $K$ and parameters $c$ and $p$. Observe that the first rate is time independent, while the second one is time-dependent. Features based on financial metrics include: moving averages (MA), MA convergence-divergence, relative strength index, real-modulated index, optimized decision index, stochastic oscillator, momentum, and pattern matching [37]. Many variants exist based on various feature transformations, including principal component analysis [41] or declustering [25,42], with no method apparently favored over others.

Figure 2 illustrates the principle of ANN-based earthquake prediction. For fully-connected feed-forward networks, the three most common approaches to input layer definition are: (*i*) $n_p$ parameters, or seismicity indicators, in a unique time window, (*ii*) $n_t$ values of one indicator over successive time windows, or (*iii*) $n_p \times n_t$ input nodes representing the evolution of $n_p$ seismicity indicators over $n_t$ time bins. (*ii*) and (*iii*) encode the temporal evolution of hypothetical precursory patterns [11,20], which is naturally built in RNNs.

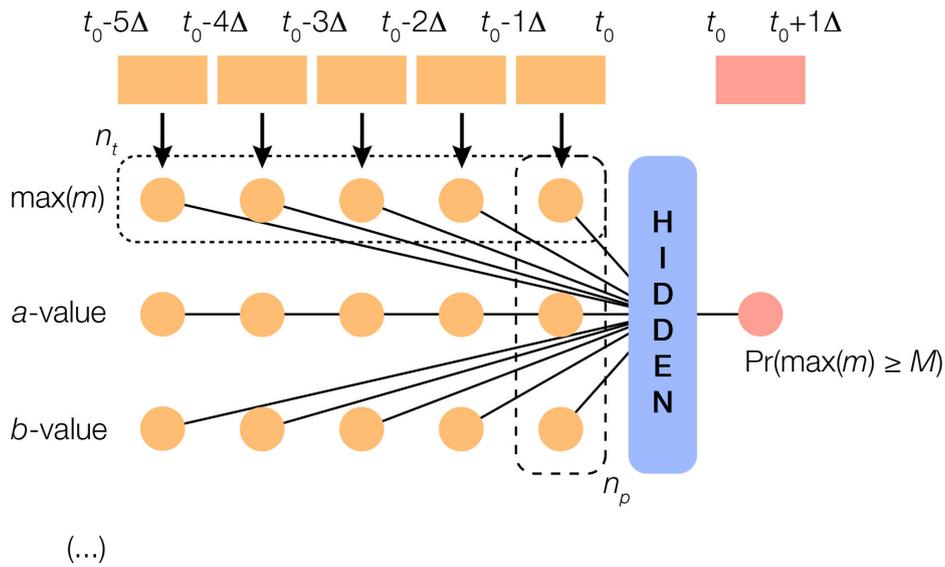

**Fig. 2** Principle of ANN-based earthquake prediction (fully-connected feed-forward case). Earthquake catalogue time series are discretized in bins from which seismicity indicators are estimated. The size of the input layer is the number of indicators $n_p$ times the number of time intervals Δ for which those parameters are estimated

Virtually all published studies part of our corpus claim positive results, with ANN models able to provide "good" earthquake predictability. The metrics considered in the corpus derive from the true positive *TP*, true negative *TN*, false positive *FP* and false negative *FN* counts of the confusion matrix. They are mainly the true positive rate *TPR* = *TP*/(*TP*+*FN*) (also known as sensitivity and recall), the true negative rate *TNR* = *TN*/(*TN*+*FP*) (also known as susceptibility) and the *R*-score, defined as *TP*/(*TP*+*FN*) - *FP*/(*TN*+*FP*) = *TPR*+*TNR*-1 (also known as True Skill Score *TSS*) [33,43]. Results vary significantly between studies but with *R*-scores greater that zero suggesting some predictive power (see section 3.1.2 for some numbers). The gain of using ANNs instead of simpler methods remains unclear since performance is only compared to a baseline in 47% of cases. Of those, only 22% use a baseline such as a Poisson null-hypothesis or randomized data. The remaining 78% mostly compare ANN results to results obtained by other machine learning methods, with the same features and data.

When the baseline is a machine learning classifier (support vector machine, decision tree ensemble, naive Bayes, *etc.*) [35, 44-45], it might not be as fine-tuned as the proposed ANN [16-17]. In other cases, the baseline can be an oversimplification of the natural (non-Poissonian) behavior of seismicity [46]. The apparent lack of proper baseline modeling will be investigated in section 3.1.

*2.2 Highly-parameterized deep learning trend*

Figure 3 highlights the progressive complexification over time of ANN models, towards deep learning, observed in the ANN-based earthquake prediction corpus. We find an increase in the number of hidden layers in fully-connected feed-forward networks (Fig. 3a), up to the extreme case of a 6-hidden layer DNN [9]. Regarding all types of ANNs, we also find a trend towards more complex architectures with Long Short-Term Memory (LSTM) networks [47] and CNNs [25] used since 2017.

We now turn our attention to the DNN of [9], which has been highly mediatized in 2018-2019, proving a recent interest in "Artificial Intelligence (AI)"-based earthquake prediction (Fig. 1) [6-7,12,48-50] (section 2.2.1). We also investigate the first use of a CNN for earthquake prediction [25] (section 2.2.2).

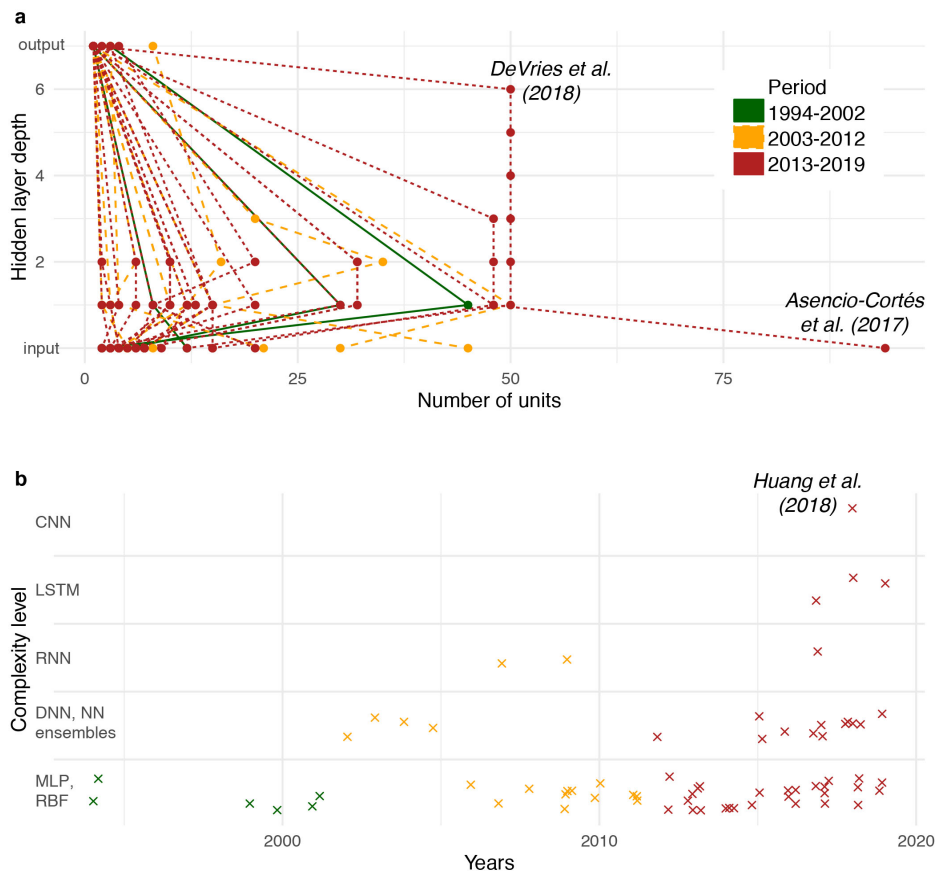

**Fig. 3** Trend towards complexification observed in the ANN-based earthquake prediction corpus: (a) Number of units per layer and number of hidden layers over three period of times; (b) Published ANNs, by complexity level, over time. Shallow ANNs (MLP, RBF) represent the simplest architectures with one hidden layer. We then consider DNNs and various ANN ensembles and other hybrids. RNNs are considered more complex than the average DNN due to the addition of the recurrent layer. LSTMs are an even more complex type of RNN. Finally, CNNs are considered the most complex architectures

2.2.1 Highlighting case studies from 2018: DeVries et al.'s DNN

  The DNN of [9] is different from previous studies in various ways. First, it did not try to predict mainshock characteristics but the spatial patterns of aftershocks (an early attempt at predicting aftershock spatial distribution had already been done by [51] but for one sequence only). Second, the authors used a global earthquake catalogue for aftershock binary classification (aftershocks present or not in geographic cells) using 12 features engineered from stress computed from mainshock rupture models, instead of seismicity indicators as previously so commonly used in the literature. Their DNN was made of 6 hidden layers, each composed of 50 nodes, yielding a total of 13,451 free parameters (Fig. 4a).

Aftershocks were defined as all events located in a fixed space-time window following a mainshock, for 199 of them worldwide. Aftershocks were then aggregated in geographic cells, labelled 1 if a cell contains at least one aftershock, 0 otherwise. The authors used also mainshock rupture data (geometry, mechanism, slip) to compute the change of elastic stress tensor $\Delta\sigma$ [52] due to a dislocation in a homogeneous 3D medium at the centroid of each cell. Finally, they defined the DNN input layer with the absolute values of the six independent components of $\Delta\sigma$, which are $|\Delta\sigma_{xx}|$, $|\Delta\sigma_{xy}|$, $|\Delta\sigma_{xz}|$, $|\Delta\sigma_{yy}|$, $|\Delta\sigma_{yz}|$, $|\Delta\sigma_{zz}|$, and their opposites $-|\Delta\sigma_{xx}|$, $-|\Delta\sigma_{xy}|$, $-|\Delta\sigma_{xz}|$, $-|\Delta\sigma_{yy}|$, $-|\Delta\sigma_{yz}|$, $-|\Delta\sigma_{zz}|$. No physical reason is behind this choice.

Based on their model input and topology, [9] obtained an Area Under the Curve (AUC) of 0.85, which appeared impressive compared to AUC = 0.58 (near-random performance) obtained for the classical Coulomb failure criterion [53], which is the main earthquake-triggering model of the current paradigm. Despite being described as a "*compact model*" in the most recent literature [6-7], Mignan & Broccardo demonstrated that similar results are obtainable with a single neuron [54]. Issues of overfitting will be discussed in section 3.1, and of lack of physical interpretability in section 3.2.

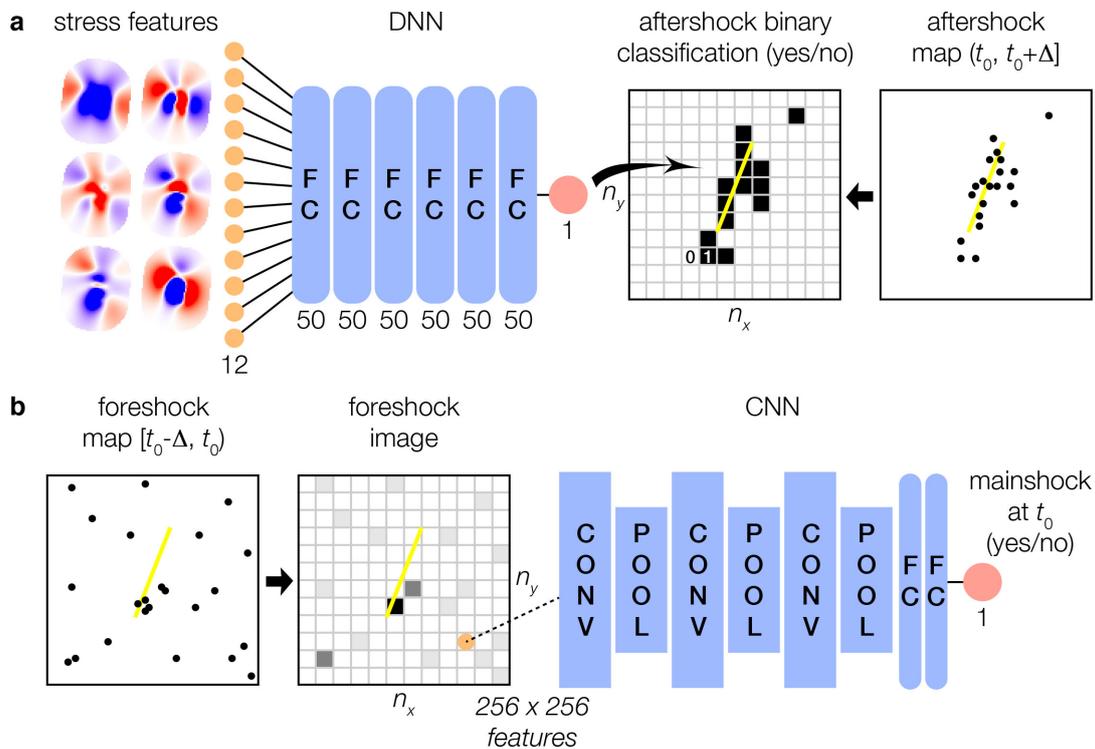

**Fig. 4** Deep learning workflows for earthquake prediction: (a) DNN of DeVries et al. [9]; (b) CNN of Huang et al. [25] - Note the different approaches, the former using 12 physics-based features, and each geographic cell of a seismicity map as one sample, to predict aftershocks in space, and the latter using 65,536 features, i.e., all the geographic cells of a seismicity map, to predict mainhocks in time. The yellow line represents the mainshock rupture, only known after the mainshock occurrence time $t_0$

2.2.2 Highlighting case studies from 2018: Huang et al.'s CNN

We should emphasize that [9] did not flatten aftershock maps as done in the computer vision applications employing fully-connected feedforward networks [55], but defined each geographic cell as one sample instead (hence leading to 131,000 samples instead of 199). This means that this study was based on structured data, with a limited number of features (12), similar to the tabulated catalogues used in previous studies (section 2.1). In contrast, Huang et al. [25] transformed Taiwanese seismicity maps into images, which led to 65,536 features (i.e., images of 256 × 256 pixels).

They fully endorsed the deep learning view of not manually designing feature vectors and letting the data speak for themselves. Their dataset was however under-sampled for such goal, with less than 500 images for training (and less than 100 for testing). They generated the images by encoding earthquake magnitude by brightness, considering all events occurring over a period of 120 days (little detail is given about minimum magnitude considered and how event magnitudes are aggregated per pixel). They labelled those images as 1 if the largest earthquake magnitude in the next 30 days was greater than or equal to 6, and as 0 otherwise. In their data preprocessing, they removed events occurring within 50 km and 7 days after the occurrence of any $M6+$ mainshock to remove predictable aftershock activity (i.e., MOL; Eq. 1b) and to focus on foreshock activity. The Huang et al. CNN is composed of three convolutional layers and three pooling layers, followed by two fully-connected layers (Fig. 4b).

Their CNN model led to an $R$-score of 0.303 while their randomized catalogues led to $R ≤ 0.065$, suggesting that the CNN captured some precursory seismicity patterns. Use of such baseline is disputable as will be discussed in the next section. Although direct comparison with other studies is difficult, some simpler ANN architectures from our corpus led to higher $R$-scores for similar magnitude thresholds ($M6+$), up to $R$-score = 0.5-1.0 [23,33]. Those high scores were however obtained on small test sets (less than 10 samples against hundreds in [25]), suggesting high uncertainties on such numbers.

## 3 Limitations of ANNs in the earthquake prediction literature
*3.1 ANN performance vs. the simplest possible baseline model*
3.1.1 Corpus investigation

The problem of overfitting is mentioned, if not addressed, in the classical ANN-based earthquake prediction literature (section 2.1). The problem of dealing with small data sets is generally emphasized [56] and the size of the ANN minimized based on a trial-and-error approach. The number of nodes in hidden layers can also be optimized via genetic algorithms [57-58], artificial bee colony [59], particle swarm optimization [60-61], *etc.* Overall, with the sample size small relative to the size of the ANN, overfitting is likely in most works present in the corpus. Overfitting is directly observable when learning curves have a sharp L shape with almost no learning done after the first epoch [62]. As we will demonstrate in section 3.1.2, under-sampling also makes performance scores much more uncertain while uncertainty bounds are virtually never given in the literature. Another problem difficult to gauge without the possibility to reproduce each study is data leakage [18].

In our corpus, the ANN is often preferred over other classifiers (decision trees, support vector machines, naive Bayes, *k*-nearest neighbor, *etc.*) based on comparisons on the same data sets [24,45]. Although we cannot easily verify the validity of those claims, it has recently been remarked that the retuning of baseline classifiers can change the ranking of what is considered a better or worse model [16-17]. Standard machine learning algorithms are performant on tabulated data made of tens of features, which is the scale of inputs observed in the ANN corpus. [63] obtained similar performances when comparing different types of ANNs with Random Forest and LPBoost ensemble. Decision tree ensembles (Random Forest and Boosting) have recently gained in interest in the earthquake prediction community [8,64], further proven by the top ranked models in a recent (lab)quake prediction competition on the Kaggle platform [65]. In an extreme case, [54] showed that a one-neuron baseline (i.e., logistic regression) is more performant than the DNN of [9] for aftershock prediction (see section 3.2).

A recently noticed gap between data scientists and seismologists [36] may explain the lack of proper baseline modelling for earthquake prediction. The most common baseline is the Poisson null-hypothesis [33,66] - although seismicity is known to cluster in both space and time [67-68]. Although some studies decluster their data in a preprocessing step [25,42], different declustering methods impact precursor analysis in different ways [69]. It is thus very difficult to establish the improvement that an ANN model provides beyond a Poisson baseline. [66] nicely showed that while their ANN model performed better than a Poisson process (for inter-arrival time prediction), it was very close to a Weibull (i.e. non-uniform Poisson) baseline. The issue of null-hypothesis testing is discussed in the Collaboratory for the Study of Earthquake Predictability (CSEP) project [70] but much less in the ANN literature. Regarding randomized data [25,58,71], it can also be an oversimplification, losing all the empirical laws of seismicity in the process, including the GR law [71].

3.1.2 Gutenberg-Richter baseline modelling

We now build a simple baseline model solely based on the GR law (Eq. 1a), and apply it to simulations of the natural behavior of seismicity, which is best described by the Epidemic-Type Aftershock Sequence (ETAS) rate model [67-68]

$$\lambda(t) = \mu + \sum_{t_i<t} K_0 e^{\alpha(m-m_c)}(t - t_i + c)^{-p} \tag{2}$$

The ETAS parameters are set to α = 2.04, $K_0$ = 0.08, $c$ = 0.011, $p$ = 1.08, and $m_c$ = 3. This model combines a stationary background seismicity term µ and an epidemic-type clustering of aftershocks, each sequence following the MOL (Eq. 1b). The process is however devoid of foreshocks that could announce the arrival of a mainshock (i.e. null-hypothesis of no precursory signal).

The model consists in "predicting the mainshock magnitude," as done in a number of the studies part of our corpus. We define a time window [$t_0$, $t_0 + \Delta$] and predict 1 if a mainshock of magnitude $m \geq m_{th}$ occurs over Δ, and 0 otherwise, with $m_{th}$ = {4, 5, 6, 7}. The outcome is the predicted rate $\lambda(\geq m_{th}) = \Delta_r 10^{a_{trained}-b_{trained}m_{th}}$ (derived from Eq. 1a), where $a_{trained}$ and $b_{trained}$ are estimated from the training period [$t_0$ - $n\Delta$, $t_0$). The parameter $\Delta_r$ = 1/$n$ represents the ratio between prediction window Δ and training window $n\Delta$, with $n$ = {1, 4, 9}.

To represent different levels of background seismic activity, we assume the $a$-value random uniform in the interval [4, 6]. For a standard $b$ = 1 (Eq. 1a), this means one background earthquake of magnitude $m = a$ on average in the time window considered, here fixed to Δ = 100 days. Testing different mainshock magnitude thresholds $m_{th}$ for various $a$-values provides a balanced data set for sound statistical analysis. Examples of simulated GR laws (with $a$ = 4, 5, or 6) are shown in Figure 5a. One example of seismicity time series is shown in Figure 5b. Results of the GR baseline model are shown in Figures 5c and 5d, for the *TPR* and *R*-score, respectively. Results critically depend on the data split $\Delta_r$ and on the magnitude threshold $m_{th}$. For 10,000 ETAS simulations realized for each ($\Delta_r$, $m_{th}$) association, we obtain max(*TPR*) = 0.96 and max(*R*-score) = 0.67. The *TPR* obviously tends to one when $\Delta_r$ is high (i.e., more chance that the prediction window mirrors the training window when $\Delta_r$ = 1) and when $m_{th}$ is small (i.e., far more chance of a small event than a larger one with $Pr(m \geq m_{th})$ saturating at 1). By definition, the *R*-score provides a more balanced result, with the best score obtained for ($\Delta_r$ = 1/4, $m_{th}$ = 5). Similar trends are observed in the ANN corpus, with the *TPR* decreasing with increasing magnitude threshold [33,72] and with the *R*-score optimal for $m_{th}$ = 5 [23]. This does not prove that the GR baseline is as performant than the published ANN models, only that such scores are obtainable with a very simple model and realistic hyperparameterizations. Any claim of improved mainshock predictability should thus be taken with caution.

Interestingly, the importance of GR and MOL features over other parameters was demonstrated in [35] but the GR law and MOL were not used as baseline models. [24] stated that their ANN was "*capable of indirectly learning* [MOL] *and GR laws.*" Despite not using an earthquake clustering baseline for reasons of simplicity, their Poisson null-hypothesis still yielded between 5 and 23% of a mainshock being predicted by chance.

It is also likely that a number of published scores are subject to high uncertainty. Taking some of the most cited works (Fig. 1), we find that [33] obtained a *R*-score of 1 for their RNN for *M*7+ earthquakes, but with only two such events in their testing set. Signs of instability are present in their summary table with *R*-score = 0.36 for *M*4.5, 0.5 for *M*5.0-5.5, 0.0 for *M*6.0 and back to 1.0 for *M*6.5. Similar patterns are observed in [23] for their probabilistic neural network with the *R*-score oscillating from 0.5 to 0.0 and back to 0.5 for *M*6.0-6.5, *M*6.5-7.0 and *M*7.0-7.5, respectively. Their test set was composed of 4, 1 and 2 mainshocks for these respective ranges. We reapplied our GR baseline model 100 times to sets of only 10 simulations and only for $m_{th}$ = {6, 7}. The maximum possible *R*-score ranges from -0.3 to 1.0 and the minimum possible *R*-score from -0.4 to 1.0. This suggests that the stochasticity of the process and the rarity of large events combined to under-sampling can lead to any possible metric result. Once again, this does not, *per se*, reject the conclusions of the published ANN-based earthquake prediction studies, only that new tests should be undertaken to validate or dismiss claims of machine learning models beating simple earthquake statistics.

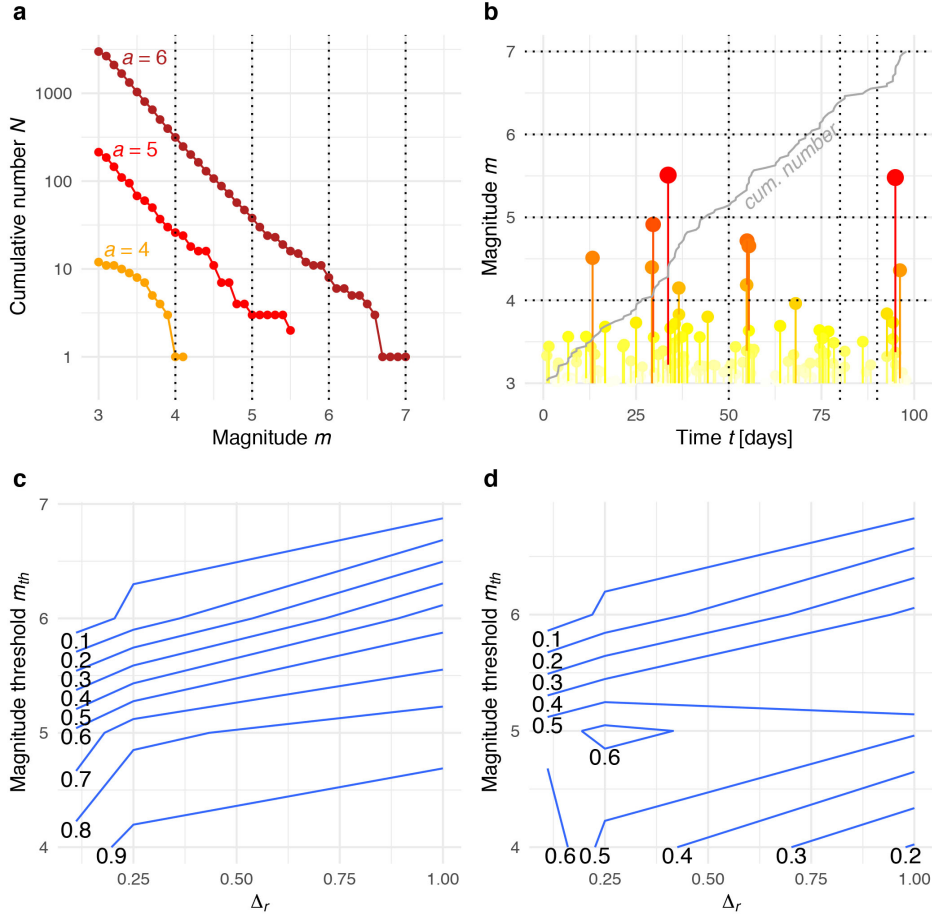

**Fig. 5** Results of a simple Gutenberg-Richter (GR) baseline model: (a) Examples of simulated GR laws for different *a*-values; (b) Example of ETAS seismicity time series. Note the natural clustering of aftershocks after any relatively high earthquake magnitude; (c) True Positive Rate (*TPR*) obtained by the baseline as a function of data split represented by $\Delta_r = 1/n$ (ratio between prediction window $\Delta$ and training window $n\Delta$) and magnitude threshold for binary classification; (d) Same as (c) but with the *R*-score

*3.2 ANN physical interpretability vs. other classifiers*

It is often accepted that defining a larger and deeper ANN does not hurt model performance. Is the complexification in ANN architecture observed in our corpus (Fig. 3) a result of overfitting or representative of truly complex patterns that only deep learning can extract? Our results of section 3.1 possibly suggest the former but it is also important to mention how different machine learning models may be (erroneously) interpreted.

Physical interpretability is critically lacking in the ANN-based earthquake prediction literature. When the black-box behavior of ANNs is not explicitly considered an "*advantage* [since] *the user need not know much about the physics of the process*" [31], it is related to complexity theory and chaos theory, and the implicit idea that a holistic system requires the use of high-variance models [73]. To the best of our knowledge, DeVries et al. [9] (section 2.2) were the first authors to seek for interpretable and meaningful physical patterns. This section therefore develops upon that study case.

A very deep ANN can be interpreted as a model of high abstraction. In computer vision, for instance, a first layer may represent simple shapes, a second layer parts of a face (such as eye, nose, ear), and a third layer, different faces [1]. When aftershock patterns are predicted by a 6-hidden-layer DNN [9], it captivates the collective imagination as to the degree of abstraction that seismicity patterns carry. This can explain a certain media euphoria about AI predicting earthquakes [12,48-50]. This is unfortunately misleading. Using the same 12 stress component features as [9], [15] demonstrated that a simpler DNN (12-8-8-1) or a shallow network (12-30-1) led to similar performances with similar prediction maps and AUC = 0.85 (Fig. 6).

In order to interpret their DNN, [9] tested various stress metrics and concluded that the sum of absolute values of independent components of $\Delta\boldsymbol{\sigma}$, $A$, the von Mises yield criterion, $\sqrt{3\Delta J_2}$, and the maximum change in shear stress, $\Delta\tau$, respectively

$$\begin{cases} A = |\Delta\sigma_{xx}| + |\Delta\sigma_{yy}| + |\Delta\sigma_{zz}| + |\Delta\sigma_{xy}| + |\Delta\sigma_{xz}| + |\Delta\sigma_{yz}| \\ \sqrt{3\Delta J_2} = \sqrt{\Delta I_1^2(\Delta\sigma') - 3\Delta I_2(\Delta\sigma')} \\ \Delta\tau = |\Delta\sigma_1 - \Delta\sigma_3|/2 \end{cases} \quad (3)$$

(where $\Delta\boldsymbol{\sigma}' = \Delta\boldsymbol{\sigma} - (\Delta\boldsymbol{\sigma}:\boldsymbol{I})/3 \cdot \boldsymbol{I}$ is the deviatoric stress change tensor with $\boldsymbol{I}$ the identity matrix; $\Delta I_1$ and $\Delta I_2$ are the 1st and 2nd invariants) yield similar AUC scores as their DNN prediction (i.e., AUC = 0.85). As such, those simple metrics were the baselines to beat. Mignan & Broccardo [54] verified that, indeed, a single neuron (i.e., logistic regression) using either of the features of Eq. (3) leads to similar performances as the published DNN (Fig. 6).

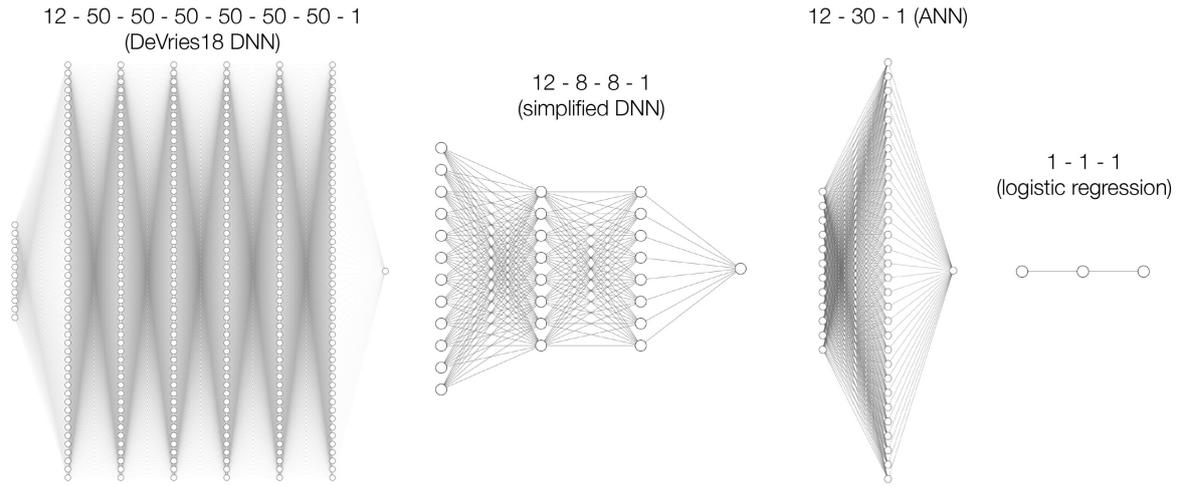

**Fig. 6** A recent example of deep learning overuse in earthquake prediction. [9] proposed a 6-hidden layer DNN to predict aftershocks in space while simpler ANN architectures suffice when using the same 12 stress features [15]. One neuron is an even better model when one stress feature is used [54]. All models led to AUC = 0.85-0.86. Figure modified from [15]. ANN topology plots generated with alexlenail.me/NN-SVG/

Figure 7a shows the aftershock patterns predicted by the original DNN with the 12 stress features ($|\Delta\sigma_{xx}|$, $|\Delta\sigma_{xy}|$, $|\Delta\sigma_{xz}|$, $|\Delta\sigma_{yy}|$, $|\Delta\sigma_{yz}|$, $|\Delta\sigma_{zz}|$, $-|\Delta\sigma_{xx}|$, $-|\Delta\sigma_{xy}|$, $-|\Delta\sigma_{xz}|$, $-|\Delta\sigma_{yy}|$, $-|\Delta\sigma_{yz}|$, $-|\Delta\sigma_{zz}|$) [9] and Figure 7b the logistic regression with single feature $A$ (Eq. 3) [54]. We see that despite the tremendous difference in model complexity, similar predictions are obtained (relating to section 3.1 on the importance of proper baseline modelling). We see also that the "unphysical" DNN feature engineering uses only absolute values of the stress components, while the "physical-based" metrics reported in Eq. 3 are convex combinations of the components of the stress tensor. Therefore, in both cases all dipolar information of the stress field is lost (see red/blue dipolar information of the stress tensor in Fig. 4a), producing—inevitably—isotropic maps[2]. What remains at first order is the distance $r$ from the mainshock rupture and a spatial scaling, which can be calibrated by the mainshock rupture displacement $d$. This is observable from Figure 7 where predicted aftershock patterns radiate from all around the mainshock rupture, where aftershocks are already known to occur.

---

[2] Any convex combination (physical or unphysical) of the stress tensor components will lose the dipolar information and therefore produce similar maps.

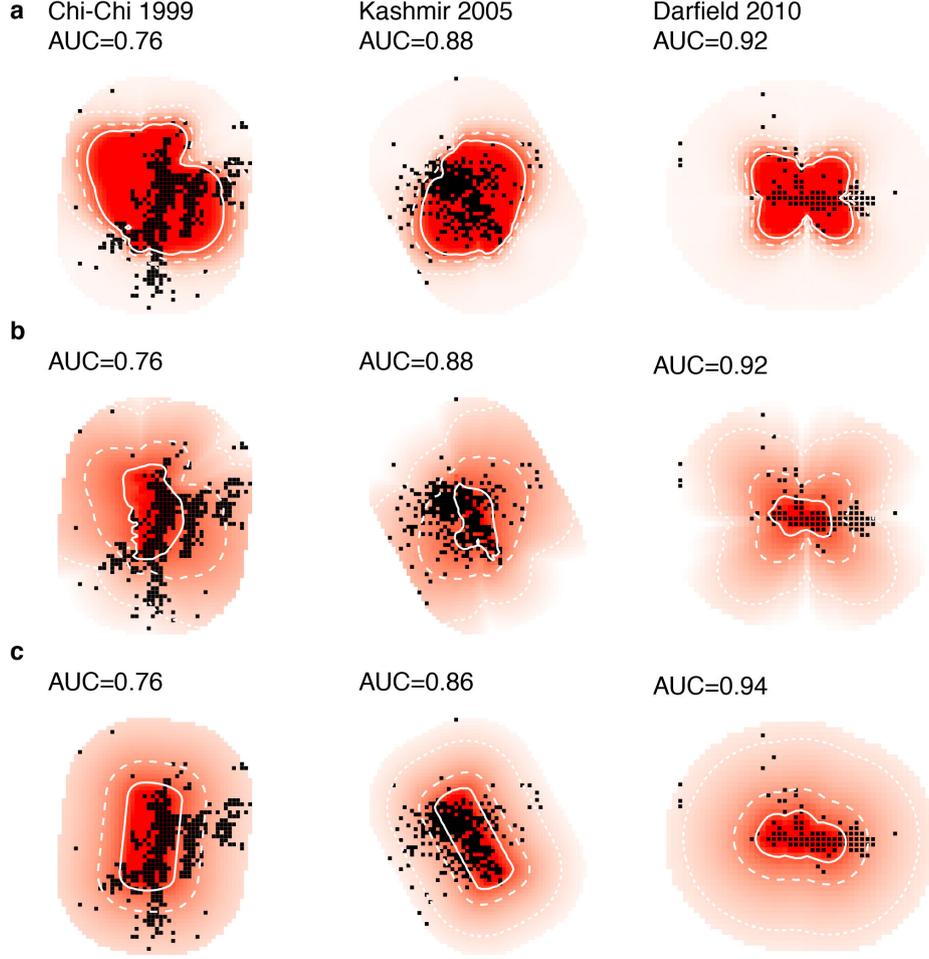

**Fig. 7** Aftershock prediction maps for three mainshocks: (a) DNN of DeVries et al. [9] with 12 features $|\Delta\sigma_{xx}|$, $|\Delta\sigma_{xy}|$, $|\Delta\sigma_{xz}|$, $|\Delta\sigma_{yy}|$, $|\Delta\sigma_{yz}|$, $|\Delta\sigma_{zz}|$, $-|\Delta\sigma_{xx}|$, $-|\Delta\sigma_{xy}|$, $-|\Delta\sigma_{xz}|$, $-|\Delta\sigma_{yy}|$, $-|\Delta\sigma_{yz}|$, $-|\Delta\sigma_{zz}|$; (b) logistic regression with unique stress feature $|\Delta\sigma_{xx}| + |\Delta\sigma_{yy}| + |\Delta\sigma_{zz}| + |\Delta\sigma_{xy}| + |\Delta\sigma_{xz}| + |\Delta\sigma_{yz}|$ [54]; (c) logistic regression with two deformation features, the distance to the rupture $r$ and the mainshock rupture displacement $d$ [54]. Dotted, dashed and solid contours represent $Pr(Y = 1) = 0.3, 0.5$ and $0.7$, respectively.

Assuming the linearized theory of elasticity, the stress-change tensor can be generally written as $\Delta\boldsymbol{\sigma} = \boldsymbol{C}\boldsymbol{\varepsilon}$ where $\boldsymbol{C}$ is the 4th order elasticity tensor (which, in the case of isotropic elasticity, has two independent constants, i.e. the Lamé parameters) and where $\boldsymbol{\varepsilon}$ is the linear strain tensor defined as the symmetric part of the displacement gradient

$$\boldsymbol{\varepsilon}(r,d) = \frac{1}{2}(\boldsymbol{\nabla u}(r,d) + \boldsymbol{\nabla u}^T(r,d)) \qquad (4)$$

where $\boldsymbol{u}(r,d)$ is the displacement field at a distance $r$ from the rupture, and $d$ is the finite rupture displacement. Following first principles, one shall thus define the features from displacement data directly, avoiding any model assumption. Having a machine learning classifier doing a mapping from deformation (mainshock geometry and kinematics) back to deformation (simplified to aftershocks occurring or not) avoids making any assumption on stress (elasticity versus poro-elasticity theory, plasticity, *etc.*), material properties (Lamé parameters), and other unknowns. In particular, this avoids having large uncertainties potentially affecting the quality of the classifier, or in other words, this avoids theoretical model bias. Recall that deformation is measurable and should be used as input layer while stress is derivative, representing subjective feature engineering (from a machine learning perspective).

Figure 7c shows the results of a logistic regression with distance to the rupture $r$ and the rupture displacement $d$ as features [54]. Parameter $r$ is defined as the minimum distance between geographic cell and the mainshock rupture plane, and $d$ as the mean slip on the rupture, from the same dataset used by [9] for stress computation (observe that both $r$ and $d$ are scalars in this version and orthogonal features). The predicted aftershock patterns are simpler and blurrier than the ones produced by [9] and generalize better. One of the main reasons is that the deep state of uncertainty is homogeneous around the fault system.

The logistic regression of [54] can be rewritten for physical interpretability as

$$Pr(Y=1|r,d) = \frac{1}{1+e^{-(\beta_0 - \beta_1 \log r + \beta_2 \log d)}} = \frac{1}{1+\hat{\beta}_0 r^{\beta_1} d^{-\beta_2}} \tag{5}$$

where $\hat{\beta}_0 = e^{-\beta_0}$. The expression is essentially a power law where $\beta_1 > 0$ controls the "geometrical attenuation" (i.e., how the probability of observing aftershock is decreasing with distance), and $\beta_2 > 0$ is controlling the "productivity" of the mainshock (i.e., how the probability of observing aftershock is increasing with rupture displacement). Such a simple parameterization of aftershock patterns is compatible with the observation that aftershocks occur closest to the mainshock rupture with their likelihood decreasing as a power-law with increasing distance [74-76], another important, yet unnamed, empirical law of Statistical Seismology.

## 4 Concluding remarks

Both the baseline we developed for mainshock magnitude prediction (section 3.1) and the logistic regression for aftershock prediction (section 3.2) provide similar results that the more complex ANNs proposed in the earthquake prediction literature that we reviewed (section 2). Those baseline models are - or can be related to - the well-known empirical laws of Statistical Seismology. We can thus conclude that ANNs so far do not seem to provide new insights into earthquake predictability, since they do not offer convincing arguments that their models surpass simple empirical laws.

This may change in the future once more features are defined and more data are available. One potential direction would be to combine different types of potential precursors (seismic, geodetic, geo-electric, *etc.*) at a global scale, which might require the flexibility of deep learning to find complex patterns. It would still be poorly guided by ANN theory, requiring a heuristic approach of trials and errors. We thus believe that first principles and strong physical constraints will remain paramount. The black box nature of an ANN and its high variance can easily lead to fallacious physical interpretations.

At the present time, deep learning is more useful and successful in unstructured seismic waveform analysis for improved event detection (e.g., DeepDetect [77], PhaseNet [78], CRED [79] or PhaseLink [80]). With those recent advances in applying CNNs and RNNs to automatically pick seismic waves, one could easily imagine applying those techniques to predict a mainshock based on foreshock seismic waves. Moreover, those techniques tend to improve the quality of earthquake catalogues by increasing the number of events ten-folds [13], so-called "Big Data catalogues". which could in turn be used as features in earthquake prediction based on structured data. A recent meta-analysis indeed showed that an increase in the amount of micro-seismicity improves precursory anomaly detection [20] but the potential of deep learning to improve earthquake forecasting remains unproven. Earthquake prediction stays the unreached Grail of Seismology.